\def\CleanCopy{1}
\documentclass[final]{opt2026}

\usepackage{booktabs}
\usepackage{array}
\usepackage{multirow}
\usepackage{xcolor}
\usepackage{placeins}
\usepackage{etoolbox}

\newif\ifhighlightchanges
\ifdefined\CleanCopy\highlightchangesfalse\else\highlightchangestrue\fi
\definecolor{changeblue}{HTML}{175DA8}
\DeclareRobustCommand{\edit}[1]{\ifhighlightchanges{\color{changeblue}#1}\else#1\fi}
\newcommand{\bibentrycolor}[1]{%
  \color{black}%
  \ifhighlightchanges
    \ifstrequal{#1}{khaled2025muonbp}{\color{changeblue}}{}%
  \fi}

\definecolor{optgreen}{HTML}{A5CF83}
\definecolor{optyellow}{HTML}{F0E76F}
\definecolor{optamber}{HTML}{ECB65F}
\definecolor{optorange}{HTML}{E89951}
\hypersetup{
  linkcolor=optorange,
  citecolor=optorange,
  urlcolor=optorange,
  pdfauthor={Mushir Akhtar, Hazem N. Nounou, Majdi Mansouri, Mohamed N. Nounou},
  pdftitle={When Does Muon Actually Win? A Paired Data-, Time-, and Tuning-Budget Benchmark},
  pdfsubject={OPT 2026: Optimization for Machine Learning},
  pdfkeywords={optimizer benchmarking, AdamW, Muon}
}

\makeatletter
\renewcommand*{\@titlefoot}{}
\makeatother

\newcommand{\adamw}{AdamW}
\newcommand{\muon}{Muon}

\title[When Does Muon Actually Win?]{When Does Muon Actually Win?\\
A Paired Data-, Time-, and Tuning-Budget Benchmark}

\makeatletter
\renewcommand*{\@email}{\small\mdseries\scshape}
\makeatother

\optauthor{%
\begin{tabular}{@{}l@{\qquad}l@{}}
\edit{\Name{Mushir Akhtar}} &
\edit{\Name{Hazem N. Nounou}}\\
\edit{\Email{mushirakhtar.ml@gmail.com}} &
\edit{\Email{hnounou@hbku.edu.qa}}\\[0.5em]
\edit{\Name{Majdi Mansouri}} &
\edit{\Name{Mohamed N. Nounou}}\\
\edit{\Email{m.mansouri@squ.edu.om}} &
\edit{\Email{mnounou@hbku.edu.qa}}
\end{tabular}}

\begin{document}

\maketitle

\begin{abstract}
\edit{Muon's gains in data efficiency motivate evaluating its performance under explicit data, time, and tuning budgets.} On a single A100 GPU, we compare a reference single-device Muon implementation with AdamW across 17.7M- and 33.6M-parameter causal language models and 2.70M-parameter vision transformers. A common 23-candidate development budget selects configurations evaluated with five matched seeds, full validation trajectories, \edit{optimizer-step timing}, and a frozen one-shot test. Muon reduces validation-loss area under the curve by 3.25--22.44\% at equal data and lowers test loss in all 20 primary pairs. \edit{In this configuration, Muon improves time-normalized loss area} by 19.68--22.31\% on vision, while AdamW is better by 3.38--6.96\% on language. \edit{The full Muon optimizer call, including auxiliary AdamW updates, consumes 56.1--69.6\% of training time, compared with 6.1--9.1\% for the AdamW baseline. Muon also has higher expected regret under restricted access to these candidate pools. These results characterize small models with reference kernels; their wall-clock ordering does not establish a ranking for larger models or optimized implementations.}
\end{abstract}

\section{Introduction}

\edit{AdamW combines coordinate-wise adaptation with decoupled weight decay and mature implementations~\citep{kingma2015adam,loshchilov2019adamw}. Muon instead forms a matrix momentum update and approximately orthogonalizes it through Newton--Schulz iteration~\citep{jordan2024muon,bernstein2024norm}. Large language-model experiments report favorable scaling and lower training loss per unit of compute~\citep{liu2025muon,khona2026emerging}.}

\edit{Efficiency can mean fewer training tokens, shorter runtime, or less tuning. These rankings can differ when optimizers have different update costs and tuning sensitivity. The original Muon implementation discussion distinguishes sample efficiency from wall-clock speed~\citep{jordan2024muon}; distributed work targets orthogonalization and communication costs~\citep{chen2026dmuon,gruntkowska2026error}.}

We pair AdamW and Muon from matched initializations on two FineWeb-Edu language models and two vision transformers trained on balanced and 100:1 long-tailed CIFAR-100. A one-seed, 23-candidate development study selects configurations for five-seed full-horizon runs. We report equal-data and equal-time validation trajectories, GPU event timing, and a frozen one-shot test evaluation. This controlled benchmark characterizes small dense models and reference kernels on one accelerator. Large-scale and distributed wall-clock rankings lie outside its scope.

Three findings organize the paper. Muon wins every primary equal-data comparison and all 20 paired test-loss comparisons. \edit{Under the measured training clock, Muon leads on vision and AdamW leads on language; the reference Muon optimizer call accounts for over half of training time.} With one randomly available candidate \edit{from each frozen pool}, expected validation-AUC regret is 13.66\% for Muon and 7.52\% for AdamW.

\paragraph{Relation to optimizer benchmarks.}
Optimizer rankings depend on search spaces, workloads, and measurement rules~\citep{choi2020empirical,schmidt2021crowded}. AlgoPerf addresses this dependence through fixed hardware, multiple workloads, time-to-result measurement, and explicit tuning protocols~\citep{dahl2023algoperf}. We apply these principles to AdamW and Muon, adding paired crossover curves, optimizer-time attribution, and post-selection test evaluation.

\section{A budget-aware comparison}
\label{sec:design}

\subsection{Workloads and optimizer routing}

The language workloads use decoder-only transformers with context length 256 on a deterministic 100M-token subset of FineWeb-Edu~\citep{penedo2024fineweb}. The micro model has 6 layers, width 256, and 17.7M parameters; the small model has 8 layers, width 384, and 33.6M parameters. Both train for 5,000 optimizer steps with an effective batch of 16 sequences. The vision workloads use a 6-layer, width-192 ViT with 4$\times$4 patches and 2.70M parameters~\citep{dosovitskiy2021vit}. One uses the 45,000-example balanced CIFAR-100 training split; the other applies exponential class-frequency decay with a 100:1 imbalance ratio and retains 9,754 examples~\citep{krizhevsky2009cifar}. Both run for 100 epochs. Each workload holds out 5,000 validation examples or 5M validation tokens during development.

For a matrix gradient $G_t$, Muon maintains momentum and applies five steps of the quintic Newton--Schulz map to a Frobenius-normalized matrix. This is a practical inexact linear minimization oracle; its degenerate case has recently been analyzed~\citep{qian2026convergence}. Writing $M_t$ for the Nesterov momentum candidate and $M_t=U\Sigma V^\top$, the resulting direction approximates $UV^\top$:
\begin{equation}
  X_{k+1}=aX_k+b(X_kX_k^\top)X_k+c(X_kX_k^\top)^2X_k,
  \quad (a,b,c)=(3.4445,-4.7750,2.0315).
  \label{eq:ns}
\end{equation}
We apply Muon to hidden attention and feed-forward matrices, including the fused QKV projection. Embeddings, normalization parameters, biases, and output heads use AdamW, following the reference routing~\citep{jordan2024muon}. \edit{Muon updates approximately 26.7\% and 42.2\% of the parameters in language-micro and language-small; auxiliary AdamW updates the remaining 73.3\% and 57.8\%. The baseline applies AdamW to every parameter.} Both methods use decoupled weight decay, gradient clipping at 1.0, 5\% warmup, cosine decay, bfloat16 arithmetic, and the same deterministic data order within a seed. Appendix~\ref{app:protocol} records the selected hyperparameters and parameter counts.

\subsection{Selection, estimands, and test isolation}

Development uses one seed and 23 candidates per workload and strategy, accumulated through an initial grid, boundary extension, and five-point audit. Selection minimizes data-normalized validation-loss AUC, breaking ties by final loss and run identifier. AdamW and Muon select interior learning rates on all workloads. Full-horizon runs use seeds 101--105; they estimate performance conditional on the selected configurations. \edit{Both data- and time-budget comparisons use these data-AUC-selected configurations. Selection stability across development seeds remains unmeasured.}

For optimizer $o$, let $L_o(x)$ denote interpolated validation loss against processed data or measured training seconds. For a paired common horizon $H$,
\begin{equation}
  \operatorname{AUC}_o(H)=\frac{1}{H}\int_0^H L_o(x)\,dx,
  \qquad
  \operatorname{gain}=1-\frac{\operatorname{AUC}_{\muon}}{\operatorname{AUC}_{\adamw}}.
  \label{eq:auc}
\end{equation}
We also report loss at equal data and equal time. Throughout the paper, gains are defined so that positive values favor the named alternative. A stable crossover is the first recorded point after which Muon remains at least 1\% below AdamW through the common data horizon. Training time excludes validation and checkpoint I/O; CUDA events measure optimizer-step time inside that interval. The seed is the replication unit. We use 10,000 paired bootstrap resamples for within-workload intervals. The pooled test-loss interval averages workloads equally and resamples seeds within workload, so its scope is the four fixed workloads rather than a population of tasks.

\edit{We froze training, selection, and all trajectory analyses before accessing the official test sets.} We evaluated each selected checkpoint once under the frozen evaluation protocol. The final set contains 40 primary evaluations (two optimizers, four workloads, and five seeds) and 24 prespecified secondary evaluations. Test outcomes do not alter any hyperparameter, checkpoint, threshold, or analysis rule.

\section{The winner depends on the budget}
\label{sec:results}

\subsection{Data efficiency is consistent; time efficiency is not}

\begin{figure}[t]
  \centering
  \includegraphics[width=0.90\textwidth]{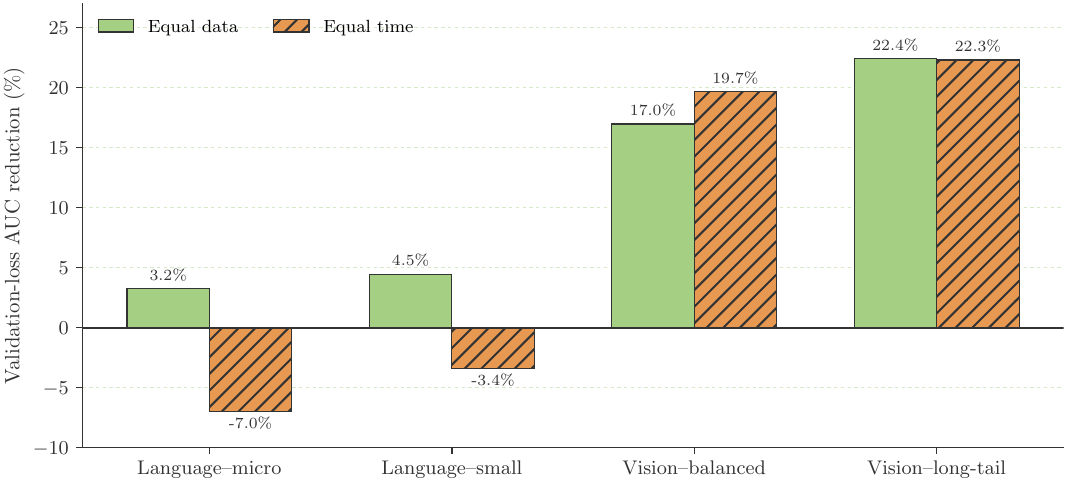}
  \caption{\edit{Mean validation-loss AUC reduction relative to AdamW over five matched seed pairs. Positive values favor Muon. Green bars compare equal data; hatched orange bars compare equal training time. With these reference kernels on one A100, the time-budget ranking reverses on both language models.}}
  \label{fig:budget}
\end{figure}

Muon lowers validation-loss AUC at equal data on every workload (Figure~\ref{fig:budget}). The gain is modest for language (3.25\% and 4.46\%) and larger for vision (16.98\% and 22.44\%). Every one of the 20 paired trajectories reaches a stable 1\%-margin crossover. Mean crossover occurs at 4.8\% and 3.0\% of the language horizons. Both vision crossovers occur at the first 1\% evaluation point, which gives an upper bound rather than a more precise crossing time.

Wall-clock accounting divides the workloads. Muon reduces time-normalized AUC by 19.68\% on balanced vision and 22.31\% on long-tailed vision. AdamW leads by 6.96\% on language-micro and 3.38\% on language-small. Equal-time endpoint loss has the same pattern: Muon is better by 39.71\% and 41.09\% on vision, while AdamW is better by 3.56\% and 0.73\% on language. The small-language endpoint interval narrowly includes zero (Appendix~\ref{app:fullresults}).

\subsection{\edit{Optimizer-step cost in the reference implementation}}

\edit{Table~\ref{tab:systems} reports the costs of the tested implementation.} Muon's throughput is 34--47\% of AdamW's with the reference kernels; optimizer steps consume 56.1--69.6\% of training time, versus 6.1--9.1\% for AdamW. \edit{The timed region includes matrix updates and auxiliary AdamW; separate Newton--Schulz timing and kernel-utilization measurements are unavailable. Peak allocated memory is 0.6--2.4\% lower for Muon, but this metric alone does not identify the cause of the slowdown. Larger token batches and optimized kernels can change the relative cost of an optimizer step~\citep{khaled2025muonbp}.}

\begin{table}[t]
\centering
\caption{\edit{Mean systems measurements over five seeds. The throughput ratio is Muon/AdamW. Optimizer shares cover the full optimizer call, including auxiliary AdamW in the reference Muon implementation.}}
\label{tab:systems}
\small
\begin{tabular}{lrrr}
\toprule
Workload & Throughput ratio & AdamW opt. share & Muon opt. share \\
\midrule
Language--micro & 0.377 & 8.8\% & 63.1\% \\
Language--small & 0.468 & 9.1\% & 56.1\% \\
Vision--balanced & 0.339 & 7.4\% & 69.6\% \\
Vision--long-tail & 0.414 & 6.1\% & 63.1\% \\
\bottomrule
\end{tabular}
\end{table}

\subsection{The data-budget advantage transfers to the test set}

\begin{figure}[t]
  \centering
  \includegraphics[width=0.86\textwidth]{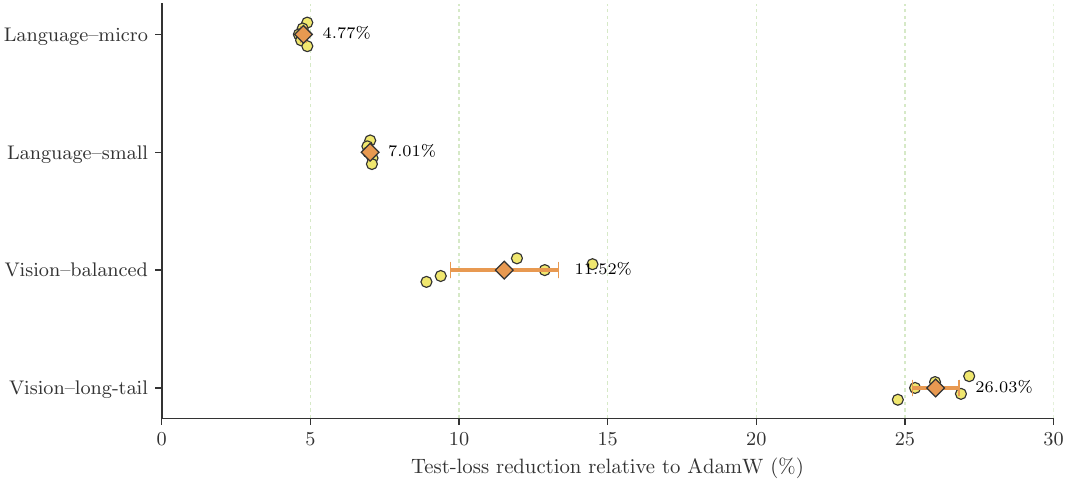}
  \caption{Paired test-loss reduction for Muon relative to AdamW. Yellow circles denote matched seeds; orange diamonds and horizontal intervals show within-workload means and 95\% bootstrap intervals. \edit{We froze all checkpoints and evaluation rules before test access.}}
  \label{fig:test}
\end{figure}

The frozen test evaluation confirms the validation result (Figure~\ref{fig:test}). Muon lowers test loss in every paired seed, with mean reductions of 4.77\% and 7.01\% on the language models and 11.52\% and 26.03\% on balanced and long-tailed vision. Language perplexity falls by 23.40\% and 32.04\%. Vision accuracy rises by 6.76 percentage points on the balanced split and 7.39 points on the long-tailed split; tail and worst-group accuracy on the long-tailed split each improve by 2.08 points. \edit{Across the four equally weighted workloads, mean relative test-loss reduction is $12.33\%$ (stratified bootstrap 95\% interval $[11.84,12.83]\%$).} This interval conditions on the chosen workloads and should not be read as uncertainty over the broader task distribution.

\section{What changes the conclusion?}
\label{sec:audits}

\begin{figure}[t]
  \centering
  \includegraphics[width=0.88\textwidth]{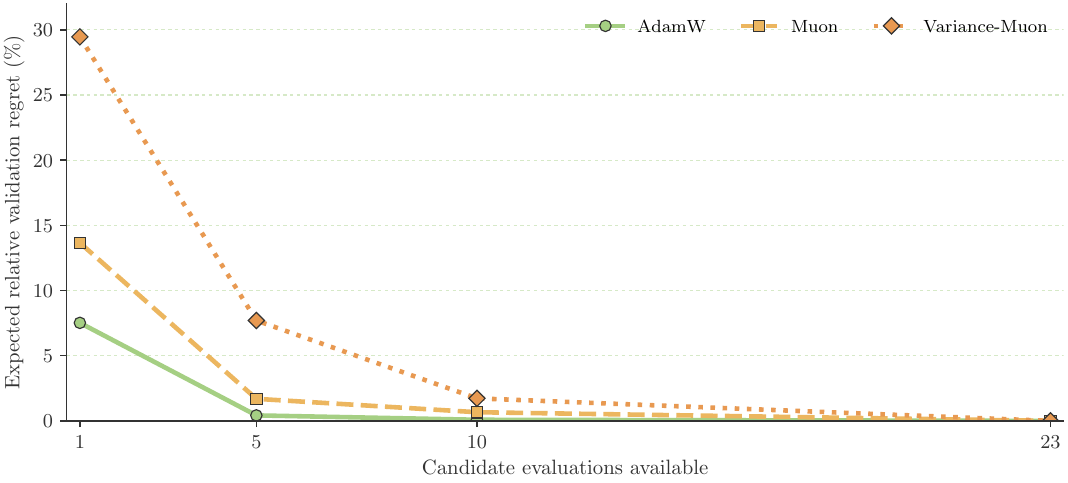}
  \vspace{0.35em}
  \small
  \begin{tabular}{rrrr}
  \toprule
  Candidates & AdamW & Muon & Variance-Muon \\
  \midrule
  1 & 7.52\% & 13.66\% & 29.46\% \\
  5 & 0.42\% & 1.69\% & 7.70\% \\
  10 & 0.09\% & 0.67\% & 1.73\% \\
  23 & 0.00\% & 0.00\% & 0.00\% \\
  \bottomrule
  \end{tabular}
  \caption{Expected relative validation-AUC regret as the candidate budget grows. Lower is better. The table gives prespecified checkpoints from the curves. \edit{Estimates average 10,000 random orderings within each workload--strategy pool, each containing 23 candidates.} This audit measures restricted access to a frozen pool, not adaptive search.}
  \label{fig:tuning}
\end{figure}

\paragraph{Tuning allowance.}
We randomly permute each 23-candidate development set 10,000 times to emulate restricted access to the frozen pools. Expected relative AUC regret with one candidate is 7.52\% for AdamW and 13.66\% for Muon; with five it is 0.42\% and 1.69\%, and with ten, 0.09\% and 0.67\%. \edit{These estimates condition on the candidate ranges, pool construction, development seed, and uniform sampling rule (Figure~\ref{fig:tuning}). Equal candidate counts do not equalize search-space coverage or tuning time. Muon's higher regret therefore describes these pools; adaptive step scaling offers a separate approach to tuning sensitivity~\citep{demidovich2026distanceaware}.}

\paragraph{Variants and approximation.}
\edit{Switching from AdamW to Muon after 25\% of training reduces test loss by 3.97--23.40\% relative to AdamW, but full-horizon Muon achieves lower loss on all four workloads. An exploratory three-seed Variance-Muon control improves both language models and balanced vision, then increases long-tailed test loss by 11.71\%. This control isolates post-orthogonalization second-moment adaptation; AdaMuon and NorMuon include additional mechanisms~\citep{si2025adamuon,li2025normuon}. In one-seed language-small audits, three rather than five Newton--Schulz steps save 21.87\% of training time with a 0.74\% AUC penalty. Restricted matrix routing saves 28--45\% with penalties of 1.13--2.76\%. Appendices~\ref{app:secondary} and~\ref{app:mechanism} report these descriptive comparisons.}

\section{Discussion and limitations}

\paragraph{A break-even interpretation.}
Let $q_o$ denote measured throughput for optimizer $o$. At time $T$, its trajectory has processed approximately $q_oT$ data units, so a time-budget comparison couples the loss curve to the implementation that produces it. Here $q_{\muon}/q_{\adamw}$ ranges from 0.339 to 0.468. The larger vision gains overcome that throughput gap; the smaller language gains do not. Faster orthogonalization raises $q_{\muon}$ and moves the break-even point even if the loss trajectory per token is unchanged. \edit{Within the frozen candidate pools, restricting the tuning budget also increases Muon's expected regret more than AdamW's.}

\paragraph{\edit{Scale and implementation.}}
\edit{The language models use widths 256/384 and 4,096 tokens per optimizer update on one A100. The relative cost of orthogonalization depends on matrix dimensions, token batch, and kernel implementation; our measurements do not isolate GPU underutilization or kernel-launch overhead. Auxiliary AdamW also covers most language-model parameters, which limits comparisons with architectures dominated by Muon-eligible matrices. Parameter count alone does not determine the resulting loss gain.}

\edit{Large-scale studies report favorable Muon compute efficiency~\citep{liu2025muon,khona2026emerging}. In an 8B model with tensor parallelism, \citet{khaled2025muonbp} report a much smaller throughput penalty and an 8\% throughput increase for MuonBP over Muon. DMuon and EF21-Muon address distributed execution and communication costs~\citep{chen2026dmuon,gruntkowska2026error}. These systems operate in different regimes; the present benchmark supports no extrapolation of its wall-clock ordering to them.}

\edit{Further limits include one-seed selection, short training horizons, crossovers resolved only at evaluation checkpoints, and fixed-pool tuning. The largest test effect, 26.03\%, comes from one 100:1 CIFAR-100 construction and cannot establish a general long-tailed result.}

\paragraph{Strength of evidence.}
\edit{The primary comparison uses five paired seeds. The three-seed variants and one-seed mechanism audits are descriptive, with less replication and narrower coverage. Broader evidence requires selection across multiple seeds, additional architectures and imbalance constructions, and optimized kernels on multiple accelerators.}

Credible comparisons should pair initializations, equalize tuning opportunities, report data- and time-normalized curves, attribute optimizer cost, and freeze test evaluation. \edit{In the four tested workloads, Muon improves data-budget performance, while the equal-time ranking remains conditional on model size, parameter routing, token batch, and implementation.}

\clearpage
\bibliography{references}

\clearpage
\appendix

\section{Experimental protocol}
\label{app:protocol}

\subsection{Data construction}

The language corpus is drawn from FineWeb-Edu, configuration \texttt{sample-10BT}, at frozen revision
\begin{center}
\small\texttt{87f09149ef4734204d70ed1d046ddc9ca3f2b8f9}.
\end{center}
Documents are assigned deterministically by a hash of the revision and stable source identifier. GPT-2 tokenization produces disjoint splits with 100M training tokens, 5M validation tokens, and 5M test tokens. The final language evaluator reads 50 deterministic batches for each checkpoint, using its frozen batch size and sequence length.

CIFAR-100 uses the original 50,000-example training split and 10,000-example test split. Fifty examples per class, selected with seed 2026, form a 5,000-example validation set. The balanced workload trains on the remaining 45,000 examples. The long-tailed workload applies exponential class-frequency decay to those training indices, from 450 examples in the largest class to 4 in the smallest, producing 9,754 examples and a 100:1 nominal imbalance ratio. The official test split is evaluated in full. Training augmentation uses random crops and horizontal flips; validation and test preprocessing are fixed.

\subsection{Models and horizons}

Table~\ref{tab:workloads} places the four architectures and training horizons on a common scale. The two language workloads vary depth, width, and parameter count; the two vision workloads hold the architecture fixed and change the training distribution.

\begin{table}[ht]
\centering
\caption{Workload definitions. The language batch is the effective number of sequences after accumulation.}
\label{tab:workloads}
\small
\begin{tabular}{lrrrrr}
\toprule
Workload & Layers & Width & Heads & Parameters & Horizon \\
\midrule
Language--micro & 6 & 256 & 8 & 17.66M & 5,000 steps \\
Language--small & 8 & 384 & 8 & 33.57M & 5,000 steps \\
Vision--balanced & 6 & 192 & 3 & 2.70M & 100 epochs \\
Vision--long-tail & 6 & 192 & 3 & 2.70M & 100 epochs \\
\bottomrule
\end{tabular}
\end{table}

Both language models use context length 256, tied input/output embeddings, pre-normalization blocks, GELU feed-forward layers of width $4d$, and no dropout. Language-micro uses batch 16 with no accumulation; language-small uses batch 8 and two accumulation steps. The ViT uses 4$\times$4 patches, six pre-normalization blocks, width 192, three attention heads, MLP ratio 4, and no dropout. Vision batch size is 128. Evaluations occur every 50 language steps and every vision epoch.

\subsection{Selected hyperparameters}

Table~\ref{tab:configs} records the configurations selected before the five-seed confirmation stage. Reporting the matrix and auxiliary learning rates separately makes Muon's parameter routing explicit.

\begin{table}[ht]
\centering
\caption{Frozen primary configurations. AdamW learning rate and weight decay apply to all parameters in the baseline. For Muon, \(\eta_M\) and \(\lambda_M\) apply to eligible matrices; auxiliary parameters use AdamW.}
\label{tab:configs}
\small
\begin{tabular}{llrrrr}
\toprule
Workload & Strategy & $\eta$ or $\eta_M$ & Weight decay & $\beta_2$ or momentum & Aux. LR \\
\midrule
Language--micro & AdamW & 0.001 & 0.00 & 0.95 & -- \\
 & Muon & 0.010 & 0.10 & 0.90 & 0.0003 \\
Language--small & AdamW & 0.0003 & 0.10 & 0.99 & -- \\
 & Muon & 0.010 & 0.10 & 0.90 & 0.0003 \\
Vision--balanced & AdamW & 0.0003 & 0.45 & 0.999 & -- \\
 & Muon & 0.010 & 0.05 & 0.85 & 0.0003 \\
Vision--long-tail & AdamW & 0.0003 & 0.45 & 0.90 & -- \\
 & Muon & 0.010 & 0.45 & 0.95 & 0.0003 \\
\bottomrule
\end{tabular}
\end{table}

Muon auxiliary weight decay is 0.10 for language and 0.05 for vision; auxiliary AdamW uses $\beta_2=0.95$. Muon has 4.72M eligible matrix parameters in language-micro, 14.16M in language-small, and 2.65M in each ViT. The corresponding auxiliary counts are 12.94M, 19.41M, and 46,180. Learning-rate schedules use 5\% linear warmup followed by cosine decay to 0.1 of the initial rate in language and 0.01 in vision.

\subsection{Tuning protocol}

The development horizon is 1,000 steps for language and 20 epochs for vision. Each of the 12 workload--strategy cells for AdamW, Muon, and the variance-adapted control contains 23 candidates. The first stage contributed 12 candidates per cell. A declared boundary extension added six, and a final audit added five low/high learning-rate and high-decay candidates. No candidate failed. The selection rule is lexicographic: normalized validation-loss AUC against data, final validation loss, then run identifier. AdamW and Muon select an interior learning rate on all four workloads. \edit{The variance-adapted control selects the lower audited learning-rate boundary and the upper weight-decay boundary in all four workloads, which motivates caution in its interpretation.}

\subsection{Timing and reproducibility}

The training clock covers forward, backward, clipping, scheduling, and optimizer steps. It pauses during validation, serialization, and logging. CUDA events surround the optimizer call, and synchronization occurs before the accumulated event durations are read. \edit{For Muon, the call includes auxiliary AdamW, matrix momentum, Newton--Schulz iterations, scaling, and decay. The reported optimizer-time share measures their combined cost.} Throughput divides cumulative tokens or examples by this training clock. Peak allocated GPU memory is the maximum value reported by PyTorch over the run.

All confirmation runs use deterministic algorithms and bfloat16 on an NVIDIA A100-SXM4-40GB. The software stack is Python 3.10.12, PyTorch 2.5.1+cu121, CUDA 12.1, and cuDNN 9.1. Checkpoint integrity is verified before each test evaluation. Test metrics are never used during training, selection, or trajectory analysis.

% \clearpage
\section{Full primary results}
\label{app:fullresults}

\begin{figure}[ht]
  \centering
  \includegraphics[width=0.92\textwidth]{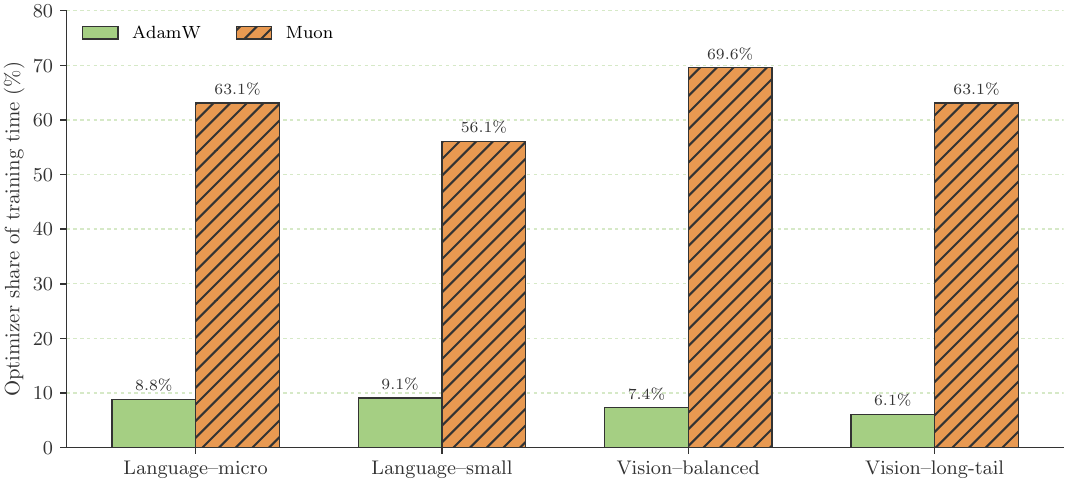}
  \caption{Optimizer-step time as a fraction of measured training time, averaged over five seeds. Green bars denote AdamW; hatched orange bars denote Muon. Validation and checkpoint I/O are excluded from the training clock.}
  \label{fig:systems-overhead}
\end{figure}

Table~\ref{tab:primary-gains} aligns validation and final-test effects under the frozen primary comparison. Table~\ref{tab:test-endpoints} gives the corresponding absolute test losses and paired intervals. The systems measurements in Table~\ref{tab:systems-full} identify the cost behind the time-budget reversal; Figure~\ref{fig:systems-overhead} makes the optimizer-time disparity visible across workloads.

\begin{table}[ht]
\centering
\caption{Primary five-seed Muon gains relative to AdamW. Positive values favor Muon. Data/time columns are validation-loss AUC gains; equal-time is the endpoint loss gain at the common time horizon.}
\label{tab:primary-gains}
\small
\begin{tabular}{lrrrr}
\toprule
Workload & Data AUC & Time AUC & Equal-time & Test loss \\
\midrule
Language--micro & 3.25\% & $-$6.96\% & $-$3.56\% & 4.77\% \\
Language--small & 4.46\% & $-$3.38\% & $-$0.73\% & 7.01\% \\
Vision--balanced & 16.98\% & 19.68\% & 39.71\% & 11.52\% \\
Vision--long-tail & 22.44\% & 22.31\% & 41.09\% & 26.03\% \\
\bottomrule
\end{tabular}
\end{table}

\edit{The language-small equal-time endpoint gain is $-0.73\%$, with a paired bootstrap 95\% interval of $[-1.60,0.22]\%$. Its time-AUC ordering therefore does not imply a resolved endpoint advantage.}

\begin{table}[ht]
\centering
\caption{Frozen final-test endpoints, mean $\pm$ sample standard deviation over five paired seeds. Intervals cover the paired relative loss reduction within each workload.}
\label{tab:test-endpoints}
\small
\begin{tabular}{lrrr}
\toprule
Workload & AdamW loss & Muon loss & Relative reduction [95\% CI] \\
\midrule
Language--micro & $5.5943\pm0.0044$ & $5.3277\pm0.0057$ & 4.77\% [4.67, 4.86] \\
Language--small & $5.5112\pm0.0085$ & $5.1250\pm0.0052$ & 7.01\% [6.95, 7.06] \\
Vision--balanced & $2.8131\pm0.0678$ & $2.4879\pm0.0347$ & 11.52\% [9.71, 13.34] \\
Vision--long-tail & $5.7715\pm0.0330$ & $4.2688\pm0.0360$ & 26.03\% [25.24, 26.82] \\
\bottomrule
\end{tabular}
\end{table}

\begin{table}[ht]
\centering
\caption{Absolute systems metrics, averaged across five seeds.}
\label{tab:systems-full}
\small
\begin{tabular}{lrrrr}
\toprule
 & \multicolumn{2}{c}{Throughput} & \multicolumn{2}{c}{Optimizer-time share} \\
Workload & AdamW & Muon & AdamW & Muon \\
\midrule
Language--micro & 116,300 & 43,879 & 8.84\% & 63.07\% \\
Language--small & 62,884 & 29,456 & 9.09\% & 56.07\% \\
Vision--balanced & 4,215 & 1,430 & 7.35\% & 69.56\% \\
Vision--long-tail & 3,137 & 1,298 & 6.09\% & 63.10\% \\
\bottomrule
\end{tabular}
\end{table}

% \FloatBarrier
\section{Tuning-budget audit}

For each workload--strategy cell, a random permutation of its 23 candidates represents an order in which configurations become available. The incumbent after budget $b$ is the minimum AUC among the first $b$ candidates. Relative regret divides its gap from the best of all 23 by the absolute best AUC. We average 10,000 permutations within a cell, then average the 12 cell curves by strategy. \edit{This estimates performance under uniform random access to the frozen pool. Pool construction includes extensions informed by earlier development outcomes, so random permutations summarize this completed pool rather than a prospective adaptive search. Different candidate ranges, sampling distributions, or development seeds can change the regret ordering.}

Figure~\ref{fig:tuning} in the main paper shows the complete budget curve and the four prespecified checkpoints. The common endpoint at 23 candidates verifies that regret is measured relative to the best configuration in the same frozen pool.

% \FloatBarrier
\section{Secondary strategies}
\label{app:secondary}

The switch strategy runs AdamW on every parameter through 25\% of the horizon, then applies Muon to eligible matrices while AdamW continues on auxiliary parameters. \edit{Development validation AUC selects the switch point from 25\%, 50\%, and 75\%.} The variance-adapted control applies an elementwise, bias-corrected second moment to the orthogonalized matrix update while preserving the same momentum, Newton--Schulz, scaling, decay, and auxiliary routing as the reference Muon implementation.

\begin{figure}[ht]
  \centering
  \includegraphics[width=0.92\textwidth]{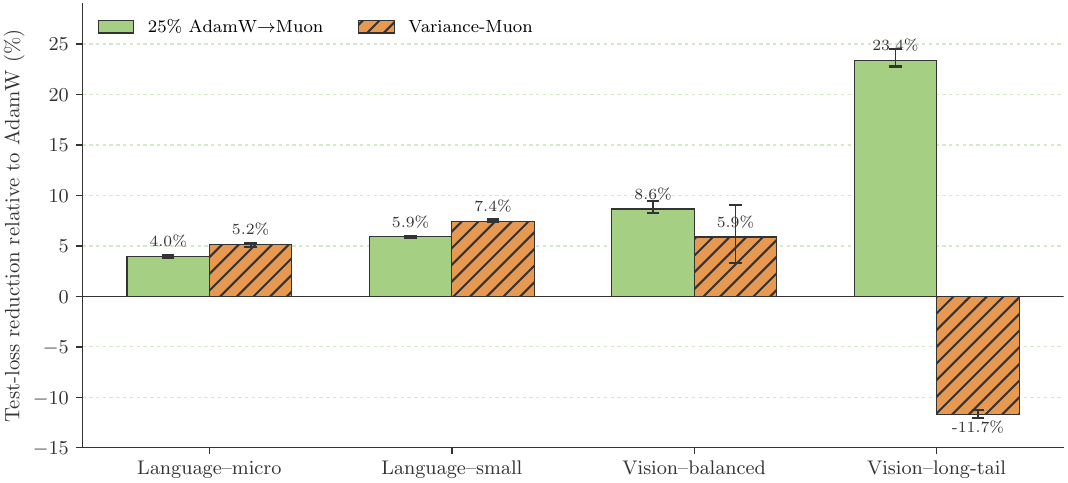}
  \caption{Three-seed final-test loss reduction relative to AdamW. Positive values favor the alternative. Green bars denote the 25\% AdamW$\rightarrow$Muon switch; hatched orange bars denote Variance-Muon. Error bars are within-workload bootstrap intervals.}
  \label{fig:secondary}
\end{figure}

Figure~\ref{fig:secondary} displays the paired uncertainty for both variants; Table~\ref{tab:secondary} provides the corresponding means. These three-seed comparisons are exploratory and are not power-matched to the five-seed primary study. The long-tailed workload separates the strategies sharply and motivates treating the variance-adapted result as a controlled ablation rather than a uniformly stronger optimizer.

\begin{table}[ht]
\centering
\caption{Secondary final-test loss reduction relative to AdamW, three paired seeds.}
\label{tab:secondary}
\small
\begin{tabular}{lrr}
\toprule
Workload & 25\% AdamW$\rightarrow$Muon & Variance-Muon \\
\midrule
Language--micro & 3.97\% & 5.16\% \\
Language--small & 5.92\% & 7.45\% \\
Vision--balanced & 8.65\% & 5.89\% \\
Vision--long-tail & 23.40\% & $-$11.71\% \\
\bottomrule
\end{tabular}
\end{table}

% \FloatBarrier
\section{Descriptive mechanism audit}
\label{app:mechanism}

The mechanism study uses the language-small development workload and seed 11. Seven learning-rate candidates are evaluated for each Newton--Schulz depth; three candidates are evaluated for each eligible-role setting. All selections use the same normalized validation-AUC rule. The role labels are: all eligible matrices; all attention projections; feed-forward up/down projections; attention output projections; and fused QKV projections. Because the model stores Q, K, and V in one matrix, the audit cannot distinguish their individual effects.

\begin{figure}[ht]
  \centering
  \includegraphics[width=0.92\textwidth]{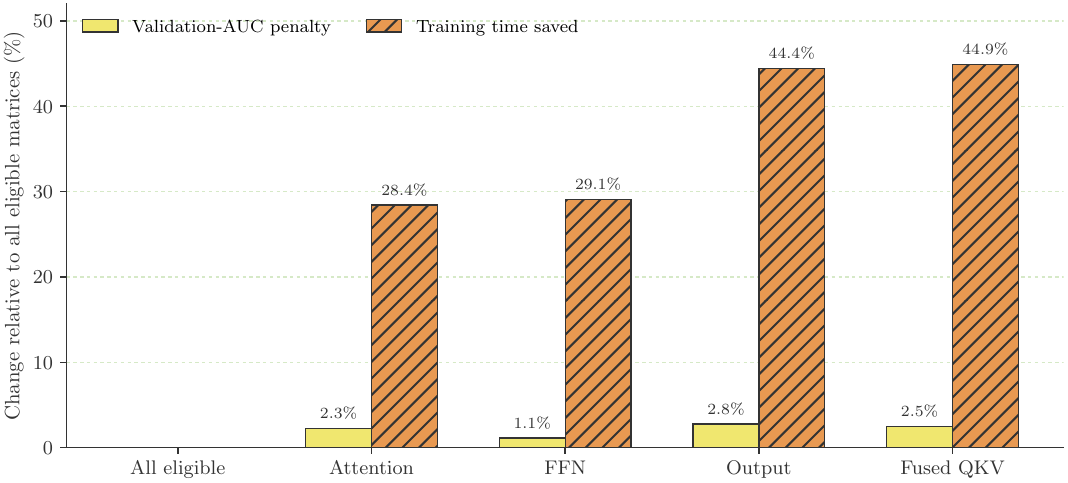}
  \caption{Validation-AUC penalty and training time saved when Muon is restricted to a subset of eligible matrices. Yellow bars show the AUC penalty; hatched orange bars show time saved relative to routing all eligible matrices through Muon. The audit uses one workload and seed and is descriptive.}
  \label{fig:mechanism}
\end{figure}

Figure~\ref{fig:mechanism} reports the quality--time trade-off from restricting eligible matrices. Table~\ref{tab:ns} isolates a second approximation axis by comparing the selected three- and five-step Newton--Schulz configurations.

\begin{table}[ht]
\centering
\caption{Selected Newton--Schulz configurations in the mechanism audit.}
\label{tab:ns}
\small
\begin{tabular}{rrrrr}
\toprule
NS steps & Matrix LR & Validation AUC & Training seconds & Orthogonality error \\
\midrule
3 & 0.01 & 6.6162 & 100.59 & 0.851 \\
5 & 0.01 & 6.5676 & 128.75 & 0.513 \\
\bottomrule
\end{tabular}
\end{table}

% \FloatBarrier
\section{Statistical scope and reporting conventions}

All primary contrasts are paired by workload and seed. Evaluation checkpoints within a trajectory are used for interpolation and integration, not as independent replicates. AUC uses trapezoidal integration after sorting and de-duplicating the budget axis; interpolation inserts a common-horizon endpoint when needed. The initial observed loss is carried to budget zero. Equal-budget horizons are the minimum attained by the paired optimizers.

Within-workload intervals resample the five paired seed effects with replacement 10,000 times using random seed 2026. The pooled final-test interval first resamples seeds separately inside each workload and then averages the four workload estimates with equal weight. It is conditional on this fixed workload set. Secondary intervals use three paired seeds. Crossover observations are interval-resolved at recorded validation checkpoints. No multiplicity correction is applied because the workload effects and resource definitions are reported as a structured profile rather than a family of interchangeable null-hypothesis tests.

All gains are direction-normalized: positive values indicate lower loss or AUC for the named alternative. Accuracy effects are percentage-point differences. Time metrics use validation curves because test evaluation occurs only after training.

\end{document}